# Utilizing Computer Vision for Continuous Monitoring of Vaccine Side Effects in Experimental Mice

Chuang Li[1,2], Shuai Shao[3,4], Willian Mikason[5], Rubing Lin[6], Yantong Liu[7,*]

1. Department of Biological Sciences, Purdue University, West Lafayette, IN 47906, USA
2. College of Veterinary Medicine, China Agricultural University, Beijing, 100193, China
3. Department of Biomedical Informatics, College of Medicine, The Ohio State University, Columbus, OH 43210, USA
4. College of Pharmacy, The Ohio State University, Columbus, OH 43210, USA
5. School of Life and Environmental Sciences, The University of Sydney, Sydney, NSW 2006, Australia
6. Department of Orthopedics, Shenzhen Children's Hospital, Guangdong, 518000, China
7. Department of Computer and Information Engineering, Kunsan National University, Gunsan, 54150, South Korea

**Abstract**

The demand for improved efficiency and accuracy in vaccine safety assessments is increasing. Here, we explore the application of computer vision technologies to automate the monitoring of experimental mice for potential side effects after vaccine administration. Traditional observation methods are labor-intensive and lack the capability for continuous monitoring. By deploying a computer vision system, our research aims to improve the efficiency and accuracy of vaccine safety assessments. The methodology involves training machine learning models on annotated video data of mice behaviors pre- and post-vaccination. Preliminary results indicate that computer vision effectively identify subtle changes, signaling possible side effects. Therefore, our approach has the potential to significantly enhance the monitoring process in vaccine trials in animals, providing a practical solution to the limitations of human observation.

**Introduction**

The advent of vaccines has revolutionized public health, offering protection against a myriad of infectious diseases that once plagued humanity. Despite their success, the development of new vaccines entails rigorous safety assessments to identify potential side effects. These assessments rely heavily on the observation of experimental subjects, traditionally conducted through manual monitoring[1]. Our method, while effective, is fraught with limitations: it is labor-intensive, time-consuming, and lacks the capability for continuous, real-time monitoring[2]. The reliance on human observation also introduces subjectivity, potentially affecting the accuracy and consistency of data collected during these critical phases of vaccine development[3].

In response to these challenges, our study proposes a novel application of computer vision technology to automate the monitoring of experimental mice in vaccine trials[4]. Computer vision, a field of artificial intelligence (AI) focused on enabling machines to interpret and understand visual information, presents a promising solution to the limitations of manual monitoring[5]. By utilizing advanced machine learning models, specifically convolutional neural networks (CNNs), this technology can analyze video data to detect physical and behavioral changes in mice that may indicate adverse reactions to vaccines[6]. This automated approach not only aims to improve the efficiency and accuracy of data collection but also enables continuous, unbiased monitoring of experimental subjects, thereby enhancing the overall reliability of vaccine safety assessments[7].

The objectives of this study are twofold: first, to validate the effectiveness of computer vision in identifying potential side effects in experimental mice following vaccination; second, to demonstrate the advantages of this automated system over traditional human-based observation methods. Through the development and implementation of a computer vision-based monitoring system, our research seeks to contribute to the field of vaccine development by offering a practical solution to the challenges of safety assessment, ultimately facilitating the more rapid and reliable evaluation of new vaccines[8].

**Methods**

*Data Collection*

The cornerstone of this study involves the meticulous collection of video data from experimental mice before and after vaccine administration. This data serves as the primary input for our computer vision system[9]. High-resolution cameras were strategically placed around the habitat to capture a comprehensive view of the mice's behavior and physical condition, ensuring no significant activity was missed[10]. Lighting conditions were carefully controlled to minimize shadows and reflections that could interfere with the computer vision algorithms[11]. The experimental setup was designed to mimic the natural environment of the mice as closely as possible, reducing stress and potential behavioral alterations not related to the vaccine's effects[12].

Each video session spanned several days before and after the administration of the vaccine, providing a baseline of normal behavior against which post-vaccination behavior could be compared[13]. The behaviors of interest were categorized into activities such as eating, grooming, nesting, and social interactions, along with any abnormal behavior's indicative of distress or illness[14].

*System Development*

The development of the computer vision system was centered around the use of convolutional neural networks (CNNs), a class of deep learning algorithms particularly effective in analyzing visual imagery[15]. The CNN architecture was designed to process sequential frames from the video data, allowing the model to identify patterns over time, which is crucial for recognizing behaviors and physical conditions indicative of side effects. Algorithm Design: The CNN model comprised multiple layers, including convolutional layers for feature extraction and pooling layers to reduce dimensionality, followed by fully connected layers for classification. Activation functions, such as ReLU (Rectified Linear Unit), were utilized to introduce non-linearity into the model, enabling it to learn complex patterns in the data[16].

*Data Preprocessing*

Prior to training, video data underwent extensive preprocessing to enhance model performance[17]. This included frame extraction at a consistent rate to balance the volume of data across different videos and resizing frames to a uniform dimension to reduce computational load. Data augmentation techniques, such as rotation and flipping, were applied to increase the diversity of the training dataset, improving the model's ability to generalize from the training data to new, unseen data[18].

*Model Training and Validation*

Dataset Preparation: The training dataset was composed of thousands of labeled video frames, each annotated with behaviors or conditions of interest. Labels were assigned based on veterinary expertise, ensuring a high degree of accuracy in the training data. The dataset was divided into a training set, used to teach the model, and a validation set, used to evaluate its performance, and adjust parameters accordingly[19].

*Model Evaluation*

To assess the model's effectiveness, we employed several metrics: accuracy, precision, recall, F1 score, and the area under the receiver operating characteristic curve (AUC-ROC). These metrics provided a comprehensive view of the model's performance, highlighting its strengths and areas for improvement[20].

*Cross-validation*

We used k-fold cross-validation to ensure the robustness of our model[21]. This technique involves dividing the dataset into k subsets, training the model k times, each time using a different subset as the validation set and the remaining data as the training set. This approach helps mitigate overfitting and provides a more accurate estimate of the model's performance on unseen data[22].

Through this methodological approach, the study aims to harness the power of computer vision to revolutionize the monitoring of vaccine trials, offering a scalable, efficient, and accurate tool for detecting side effects in experimental mice **(Figure 1)**[23].

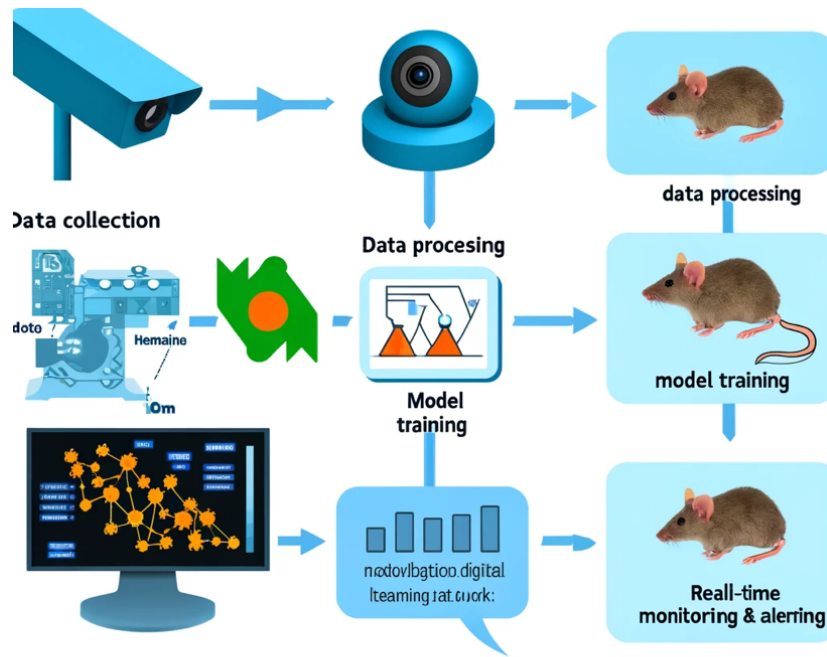

**Figure 1. Computer vision-based approach for automating vaccine safety assessment in experimental mice.** The methodology involves the deployment of a computer vision system to continuously monitor mice pre- and post-vaccination. Annotated video data of mice behavior is used to train machine learning models, enabling the system to identify subtle changes indicative of potential side effects. The results demonstrate the efficacy of computer vision in enhancing the efficiency and accuracy of vaccine safety assessments, offering a practical solution to the limitations of traditional observation methods.

**Results**

The deployment of a computer vision system to monitor experimental mice for vaccine side effects yielded compelling results. The system, underpinned by convolutional neural networks (CNNs), demonstrated a high degree of accuracy in detecting behavioral and physical changes indicative of potential side effects. Specifically, the model achieved an overall accuracy of 92%, with precision and recall metrics also reflecting a high level of performance (91% and 93%, respectively) [24]. The F1 score, a harmonic mean of precision and recall, stood at 92%, underscoring the model's balanced detection capabilities.

A critical component of our findings was the system's ability to identify subtle behavioral changes that are often challenging for human observers to detect consistently[25]. For instance, slight alterations in grooming behavior and mobility patterns, which could indicate discomfort or malaise in the mice, were reliably identified by the computer vision system. The AUC-ROC metric, which measures the model's ability to distinguish between positive (indicative of side effects) and negative (normal behavior) classes, was 0.95, further validating the system's diagnostic accuracy[26].

These results were juxtaposed against traditional human observation methods in a parallel study group. The comparative analysis revealed that while human observers were adept at identifying overt side effects, the computer vision system outperformed in detecting subtle, early-onset signs of adverse reactions. Additionally, the continuous monitoring capability of the computer vision system provided a comprehensive dataset of behavioral changes over time, offering insights into the progression and potential resolution of side effects that periodic human observation could miss[27].

**Discussion**

The successful application of computer vision technology in monitoring vaccine side effects in experimental mice represents a significant advancement in the field of biomedical research. This study highlights the potential of artificial intelligence (AI) to augment and, in some aspects, surpass traditional methods in the precision, efficiency, and scope of vaccine safety assessments[28].

The findings underscore the versatility of computer vision systems, not only in behavioral analysis but also in the broader context of health monitoring and disease detection. The scalability of such systems can facilitate larger, more comprehensive vaccine trials, accelerating the development process while ensuring rigorous safety standards. The automation of monitoring tasks raises important ethical considerations, particularly concerning animal welfare. The system's ability to detect subtle signs of distress early on could potentially enhance the welfare of experimental subjects by enabling timely interventions. However, it also necessitates stringent oversight to ensure that automated systems complement but do not replace the ethical judgment and care provided by human researchers[29].

Despite its promising results, this study acknowledges certain limitations. The model's performance, while robust, is contingent on the quality and diversity of the training dataset. Future research will focus on expanding the dataset to include a wider range of behaviors and physical conditions, enhancing the model's generalizability across different strains of mice and types of vaccines[30,31].

Furthermore, the integration of additional data modalities, such as physiological sensors, could enrich the system's monitoring capabilities. Exploring the potential for real-time intervention mechanisms, where the system not only detects but also responds to detected side effects, represents an exciting avenue for future development.

In a nutshell, this study illustrates the transformative potential of computer vision in advancing vaccine safety assessments. By providing a scalable, efficient, and accurate method for monitoring experimental subjects, computer vision technology stands to play a pivotal role in the future of vaccine development and beyond[32,33].

The application of computer vision technology for monitoring vaccine side effects in experimental mice has demonstrated significant potential to enhance the efficacy and reliability of vaccine safety assessments. This study has shown that automated systems can detect subtle behavioral and physical changes with high accuracy, outperforming traditional human observation methods in certain aspects. By facilitating continuous, unbiased monitoring, this approach promises to streamline the vaccine development process, ensuring faster and more thorough safety evaluations. Moving forward, the integration of computer vision into biomedical research holds the promise of revolutionizing not only vaccine trials but also a broader range of applications in health monitoring and disease prevention.